\documentclass[10pt,twocolumn,letterpaper]{article}

\usepackage{iccv}
\usepackage{times}
\usepackage{epsfig}
\usepackage{graphicx}
\usepackage{amsmath}
\usepackage{amssymb}
\usepackage{cite}
\usepackage{multirow}
\usepackage{color}

\usepackage[pagebackref=true,breaklinks=true,letterpaper=true,colorlinks,bookmarks=false]{hyperref}

\iccvfinalcopy 

\def\iccvPaperID{} 

\ificcvfinal\fi
\begin{document}


\title{Convolutional Channel Features}
\author{Bin Yang\qquad Junjie Yan\qquad Zhen Lei\qquad Stan Z. Li\\
Center for Biometrics and Security Research \& National Laboratory of Pattern Recognition\\
Institute of Automation, Chinese Academy of Sciences, China\\
{\tt\small \{bin.yang, jjyan, zlei, szli\}@nlpr.ia.ac.cn}}
\date{}
\maketitle

\begin{abstract}
Deep learning methods are powerful tools but often suffer from expensive computation and limited flexibility. An alternative is to combine light-weight models with deep representations. As successful cases exist in several visual problems, a unified framework is absent. In this paper, we revisit two widely used approaches in computer vision, namely filtered channel features and Convolutional Neural Networks (CNN), and absorb merits from both by proposing an integrated method called Convolutional Channel Features (CCF). CCF transfers \emph{low-level} features from pre-trained CNN models to feed the boosting forest model. With the combination of CNN features and boosting forest, CCF benefits from the richer capacity in feature representation compared with channel features, as well as lower cost in computation and storage compared with end-to-end CNN methods. We show that CCF serves as a good way of tailoring pre-trained CNN models to diverse tasks without fine-tuning the whole network to each task by achieving state-of-the-art performances in pedestrian detection, face detection, edge detection and object proposal generation.
\end{abstract}

\section{Introduction}

Many object detection solutions can be taken as the combinations of the feature extractor and classifier. Combinations such as Haar-like feature with boosting \cite{viola2004robust}, HOG feature with SVM \cite{dalal2005histograms}, multiple channel features with boosting \cite{dollar2014fast} and fine-tuned high-level CNN features with SVM \cite{girshick2014rich} have largely improved the object detection. Among these combinations, the multiple channel features with boosting and the fine-tuned high-level CNN features with SVM \cite{girshick2014rich} show the most promising performances on various detection tasks. 

The multiple channel features approach can be seen as an improved version of the Viola-Jones framework \cite{viola2004robust} with carefully hand-crafted channel feature representation and more sophisticated boosting algorithm. Multiple channel features first showed great performance in pedestrian detection \cite{ICF}, and was later generalized to face detection \cite{yang2014aggregate}, edge detection \cite{dollar2014fastedge} and object proposal generation \cite{zitnick2014edge}. Recent improvements are gained by applying learned or sophisticated hand-crafted filters on the HOG+LUV channels \cite{nam2014local,Zhang2015Cvpr}. Besides the accuracy, it typically runs at real-time speed and has very few parameters. Current bottleneck mainly lies in the representation capacity of the hand-crafted feature representation since the performance saturates as we add more and more hand-crafted filters (from tens to hundreds then to thousands) on the channel features.

The fine-tuned high-level CNN feature with SVM has recently shown extreme power \cite{girshick2014rich,oquab2014learning} in challenging tasks. Typically, a CNN model previously trained on ImageNet classification task is fine-tuned for new task and then the mid-level or high-level features of the CNN are extracted and fed into a SVM classifier. Owing to its large improvements in image classification, the learned feature hierarchies also set up new records in other vision tasks, like object detection \cite{girshick2014rich}, semantic segmentation \cite{gupta2014learning} and fine-grained category detection \cite{zhang2014part}. The main advantage of this kind of approach is that the CNN has large capacity to handle large-scale training data and the learning process is end-to-end. However, currently CNN is often accompanied by huge computation complexity in inference and learning, and the model size is usually large (e.g., more than 500MB for widely used VGG net \cite{simonyan2014very}).

In practice, we always desire for better performance and lower computation/storage cost. This motivates us to build a bridge between the above two approaches and gain benefits from their respective advantages at the same time. Specifically, we extend the multiple channel features to low-level feature maps transferred from a CNN model trained on ImageNet image classification task, or equivalently speaking, replace the high-level connections in CNN with a boosting forest. The advantages are twofold: the transferred feature maps from CNN improve the representative capacity in channel features, while the boosting forest absolves the painstaking fine-tuning of high-level connections in CNN during the adaptation to various classification/regression problems. We name the new method as Convolutional Channel Features (CCF). We explore multiple design choices concerning CCF and prove the representative capability of low-level features in CNN. We also introduce techniques to accelerate the feature pyramid computation like feature maps approximation at nearby pyramid scales and shared convolution through patchwork \cite{dubout2012exact,iandola2014densenet,papandreou2014untangling}, making CCF more computationally efficient under the sliding window setting.

The contributions of the paper are multifold:

1. From the angle of channel features, CCF extends the conventional channel features approach from HOG+LUV based features to convolutional feature maps, bringing performance gain at the cost of limited increase in computation, showing promises of learning better representation (instead of hand-crafting) in channel features approach.

2. From the CNN perspective, CCF shows that: 1) high-level connections in conventional CNN models, like convolutional and fully-connected layers, can be discarded and replaced with a boosting forest model learned with respect to different tasks, resulting in a more efficient algorithm without loss in accuracy; 2) the low-level feature representation in pre-trained CNN feature hierarchies can generalize well to diverse tasks, ranging from edge detection to pedestrian detection, without even fine-tuning to each domain.

3. CCF sets new records in 3 out of 4 vision tasks (not all metrics in edge detection), beating not only channel features variants, but also CNN based methods which either fine-tune pre-trained models to new tasks or delicately design special architecture/loss. Codes are released\footnote{\url{https://bitbucket.org/binyangderek/ccf}} for reproduction of CCF.

The remainder of the paper is organized as follows. The related work is reviewed in Section 2. In Section 3 we present the proposed method along with the feature choices, feature computation and model learning. We give analysis of the representation and experimental results on diverse tasks in Section 4 and conclude the paper in Section 5.
\section{Related work}

This work relates closely with two fundamental computer vision models, the framework of channel features~\cite{ICF}, and the Convolutional Neural Networks~\cite{krizhevsky2012imagenet,lecun1998gradient}. The channel features approach, as one of the most influential subsequences of the seminal Viola and Jones framework~\cite{viola2004robust}, has been earning its reputation with the outstandingly effective and efficient HOG+LUV channel-wise feature representation, since the successful debut in pedestrian detection. Given that feature representation, a decision tree model learned in a boosting manner can achieve state-of-the-art performance in visual recognition task. Another important advantage is that it is very efficient, as the computation processes of both feature extraction and model inference cost little. Besides visual recognition task (e.g. pedestrian detection), channel features are also successfully applied to other vision tasks like pose estimation~\cite{DollarCVPR10pose}, edge detection~\cite{dollar2014fastedge} and object proposal generation~\cite{ZitnickDollarECCV14edgeBoxes}. Recently, variants of channel features have made great efforts in improving it, like Roerei~\cite{benenson2013seeking}, LDCF~\cite{nam2014local}, SquaresChnFtrs~\cite{benenson2013seeking}, InformedHaar~\cite{zhang2014informed} and Checkerboards~\cite{Zhang2015Cvpr}. One common contribution of these is to apply additional filters on the HOG+LUV channels to increase the capacity of the feature~\cite{Zhang2015Cvpr}. The filters can be in varying forms, like simply sum-pooling~\cite{benenson2013seeking}, unsupervised PCA filters~\cite{nam2014local}, or Haar-like filters~\cite{zhang2014informed}. In an investigation of  progress in pedestrian detection~\cite{Benenson2014Eccvw}, it is pointed out that recent improvements in pedestrian detection are mainly gained from better feature representation. However, as the number of hand-crafted filters rises, the performance saturates.

In the other camp, the Convolutional Neural Networks have brought a revolution to the computer vision community in recent years. Two features of CNN models play the key role in its success: 1) The feature representation is learned in a hierarchical way, making it more representative. With increasing depth of layers, the representative capacity is enhanced, making it capable to handle recognition task of hundreds or thousands of categories. 2) CNN features have a good generalization ability. Specifically, \cite{donahue2013decaf} shows that high-level CNN features can be transferred to generic recognition tasks without fine-tuning. \cite{oquab2014learning} further proves that fine-tuning improves the performance. Through fine-tuning, CNN features learned on large-scale recognition tasks have been successfully transferred to various vision tasks like object detection~\cite{girshick2014rich}, semantic segmentation~\cite{long2014fully} and action detection~\cite{gkioxari2014r}, but they suffer from heavy computation cost caused by high-level connections \cite{donahue2013decaf,girshick2015fast}.

As the above two models are developed individually, we want to build a bridge between them, by replacing the hand-crafted channel features with low-level CNN features. There have already been some works attempting to connect CNN with popular shallow models, like DPM~\cite{savalle8deformable,girshick2014deformable} and Regionlet~\cite{zou2014generic}. They both use the output of the last convolutional layer as feature representation and concatenate the CNN features with another structural model. There is also work that harnesses low-level CNN features, like \cite{karianakis2015boosting} that uses the output of first convolutional layer to learn a two-class object classifier for proposal generation, then the RCNN approach is used for object detection. Our work differs from these in that: 1) we use low-level CNN features as a universal representation (without fine-tuning) for vision tasks ranging from edge detection to pedestrian detection; 2) we apply boosting forest model directly on the feature maps to solve vision problems, which owns the advantages of fewer parameters, smaller model size and faster inference, compared with conventional CNN methods.

\begin{figure}[t]
\begin{center}
   \includegraphics[width=0.9\linewidth]{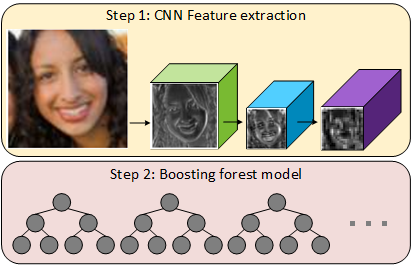}
\end{center}
   \caption{The pipeline of Convolutional Channel Features (CCF), which is a concatenation of low-level CNN feature extraction and boosting forest model learned with respect to diverse tasks.}
\label{fig:pipeline}
\end{figure}

\section{Proposed method}

As shown in Fig. \ref{fig:pipeline}, CCF is composed by concatenating two individual components, the CNN feature extraction part and the boosting forest part. The CNN feature extraction part extracts feature representation, using only the first few layers from a pre-trained CNN model. The boosting forest part learns different models for different tasks, with each node in decision trees dependent on one pixel value in the given feature maps.

In the following part of this section, we introduce the selection of CNN features used for CCF, techniques to accelerate the feature extraction process under the sliding window setting and how to learn boosting forest models for different vision tasks.

\begin{table}
\begin{center}
\newcommand{\tabincell}[2]{\begin{tabular}{@{}#1@{}}#2\end{tabular}}
\begin{tabular}{|l|c|c|c|c|c|}
\hline
 & \tabincell{c}{Output\\layer} & \tabincell{c}{\#Output\\maps} & \tabincell{c}{Filter\\size} & \tabincell{c}{\#Ds} & \tabincell{c}{Miss\\Rate(\%)}\\
\hline\hline
ACF & - & 10 & 3 & 4 & \textbf{41.22}\\
LDCF & - & 40 & 7 & 4 & \textbf{38.66}\\
\hline
\multirow{5}{*}{ANet-s1} & conv1 & 96 & 11 & 4 & 61.65\\
 & conv2 & 256 & 5 & 4 & 51.52\\
 & conv3 & 384 & 3 & 4 & \textbf{43.73}\\
 & conv4 & 384 & 3 & 4 & 48.37\\
 & conv5 & 256 & 3 & 4 & 53.37\\
\hline
\multirow{4}{*}{VGG-16} & conv2-2 & 128 & 3 & 4 & 53.86\\
 & conv3-3 & 256 & 3 & 4 & 31.28\\
 & conv4-3 & 512 & 3 & 8 & \textbf{27.66}\\
 & conv5-3 & 512 & 3 & 16 & 51.52\\
\hline
\multirow{4}{*}{VGG-19} & conv2-2 & 128 & 3 & 4 & 51.25\\
 & conv3-4 & 256 & 3 & 4 & 33.56\\
 & conv4-4 & 512 & 3 & 8 & \textbf{30.17}\\
 & conv5-4 & 512 & 3 & 16 & 55.55\\
\hline
\multirow{4}{*}{GNet} & conv2 & 192 & 3 & 4 & 45.06\\
 & icp1 & 256 & - & 8 & 38.44\\
 & icp2 & 480 & - & 8 & \textbf{31.66}\\
 & icp3 & 512 & - & 16 & 35.99\\
\hline
\multirow{4}{*}{GNet-s1} & conv2 & 192 & 3 & 4 & 49.39\\
 & icp1 & 256 & - & 4 & 41.85\\
 & icp2 & 480 & - & 4 & \textbf{32.18}\\
 & icp3 & 512 & - & 8 & 32.87\\
\hline
\end{tabular}
\end{center}
\caption{Comparison of different feature choices validated on a small train/test split of Caltech Pedestrian Benchmark training set. `\#Ds' means the down-sampling ratio between the output feature maps and the input image patch. `-s1' means that we change the application stride of filters in the first convolutional layer to 1, and only `GNet' in the table has a different stride of 2 among all nets.}
\label{tab:design}
\end{table}

\subsection{Selection of feature representation}

We compare various feature choices among several popular CNN models, which are AlexNet (`ANet' for short)~\cite{krizhevsky2012imagenet}, VGG net~\cite{simonyan2014very} and GoogLeNet (`GNet' for short)~\cite{szegedy2014going}. The performance is evaluated through a standardized evaluation protocol that we personally defined on Caltech pedestrian benchmark~\cite{Dollar2012PAMI}. Specifically, we use set00-04 as training set and set05 as test set. All images are sampled from the videos in the benchmark at an interval of 20 frames for the sake of efficiency. For comparison, we train an Aggregate Channel Features (ACF~\cite{dollar2014fast}) model and a Locally Decorrelated Channel Features (LDCF~\cite{nam2014local}) model as two baselines using the open source toolbox \cite{PMT}. We adopt hard negative mining strategy in training baseline models, and all collected negative samples by ACF model are stored for all experiments of CNN feature selection for the sake of training speed. Note that this has an impact on performances but the difference is negligible for the purpose of feature selection. The model parameters for each case are fixed as well. Specifically, the window size is set to $128\times64$, and 2048 depth-3 decision trees are learned with RealBoost algorithm. There's one exception which is the down-sampling factor. In multiple channel features, it has been proven~\cite{ICF} that a down-sampling factor of 4 performs best. In our case, as different layers in CNN often have different number of pooling layers under them, we set a minimum down-sampling of 4 for all cases. That is to say, if the pooling factor of the extracted layer is smaller than 4, we add additional average pooling to the feature maps in order to avoid feature representation with very large feature dimension. It should be noted that the convolution stride of first convolutional layer in `ANet' and `GNet' is changed to 1 in our case (marked as `ANet-s1' and `GNet-s1', `GNet' has a stride of 2), since now we are extracting feature representation on an image patch that is smaller than the $224\times224$ whole image.

In Table~\ref{tab:design}, we present results of various feature selections. The performance is measured by Log-average Miss Rate under the reasonable setting~\cite{Dollar2012PAMI} (pedestrians taller than 50 pixels). We choose recently proposed large (deep) CNN models trained on ImageNet dataset as feature candidates, as these models outperform small models trained on small datasets remarkably in our preliminary experiments. Deep models also have the advantage of simple image pre-processing which is just mean extraction, compared with global contrast normalization and ZCA whitening as used in NIN model \cite{lin2013nin}. We bold the best performance in each entry, among which a $\sim10\%$ improvement over the ACF and LDCF baselines can be seen. We first clarify the effect caused by changing the stride of first convolutional layer. By comparing between `GNet' and `GNet-s1', the change of stride leads to a negligible drop in performance. Based on this, our first observation is that large filter size acts poorly in channel features framework, by taking `ANet' as an example. One possible reason is that larger filter loses focus on local cues like edges, making it more appropriate for representing whole images rather than small patches. This observation also stands in channel features, where large local scale diminishes the performance greatly \cite{ICF}. The second observation is that the best choice in each entry is similar, i.e., around the convolutional layer whose accumulated pooling factor is 4 or 8. There are two factors affecting this observation, the layer depth and the down-sampling factor. Since channel features are general-purposed and dense feature representation, it is reasonable to prefer low-level features in CNN to the task-specific, sparse high-level one.

\begin{figure}[t]
\begin{center}
   \includegraphics[width=0.95\linewidth]{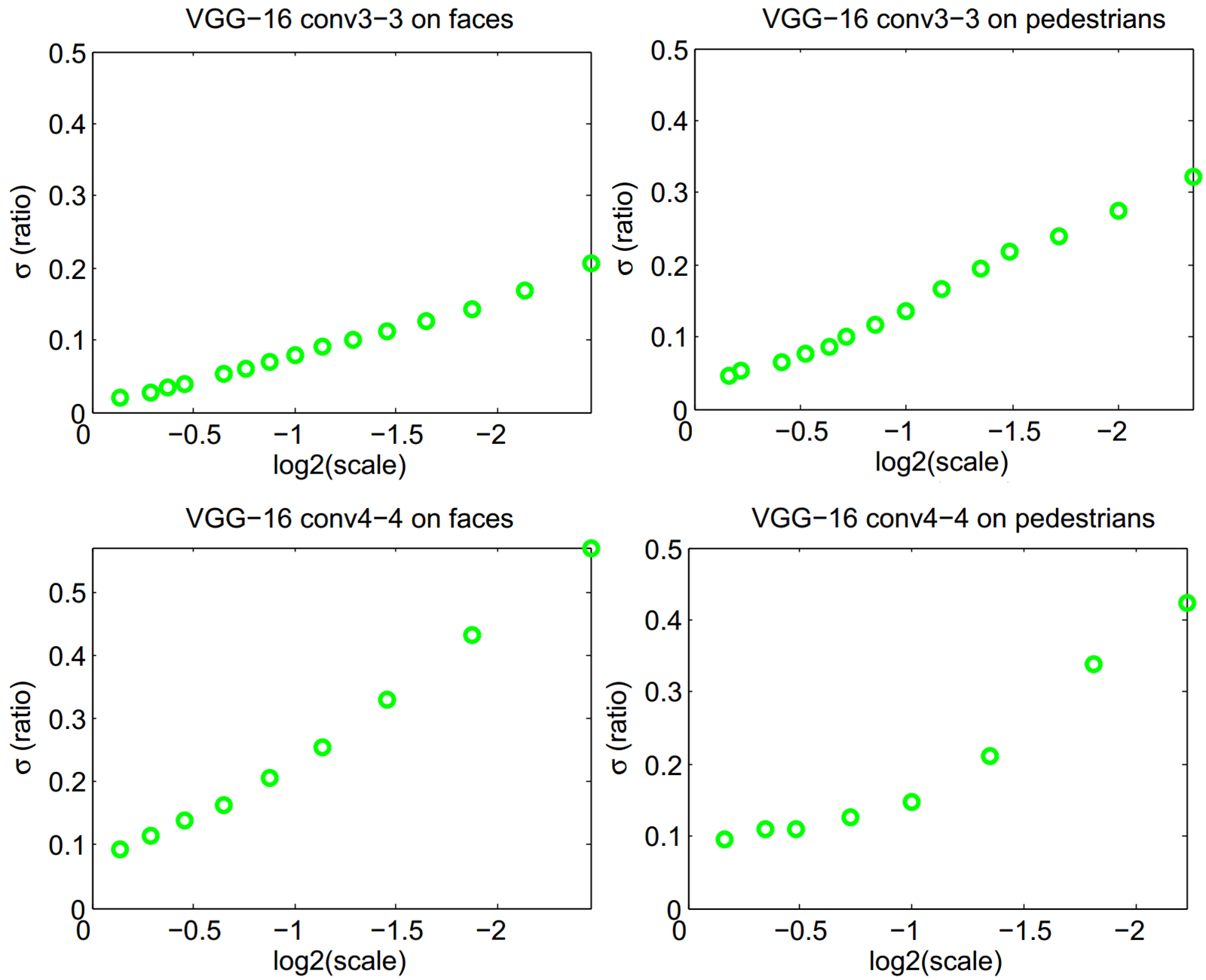}
\end{center}
   \caption{Power law fitting results of CCF at different scales on 100 randomly selected images.}
\label{fig:powerlaw}
\end{figure}
 
\subsection{Acceleration in feature pyramid computation}

CNN based object detectors usually use object proposals as input \cite{girshick2014rich} for better performance in speed (no need for multi-scale pyramid construction) and accuracy (fewer false positives), while CCF is naturally applicable to this approach, we use sliding window mechanism instead in order to: 1) compare with variants of filtered channel features \cite{Zhang2015Cvpr} fairly with fixed settings; 2) reveal the raw performance of CCF without resorting to other approach. While pyramid construction of channel features can be quite efficient, the same process of CNN features can be expensive due to restriction of a fixed input size of CNN models. In the following we discuss two known techniques used in accelerating pyramid construction and see how they can be applied to CCF. In our case, we observe a $4\sim6$ times speedup with the acceleration techniques implemented.

\textbf{Power law in scales:} The definition of power law in scales is that feature responses of specific feature types on natural images at different scales are subject to the power law, therefore when computing the feature pyramid of an image, we can use the feature representation at one scale to approximate the feature representations at nearby scales instead of iteratively computing each scale \cite{dollar2014fast}. To this end, we wonder whether it holds in the circumstance of CCF. Fig. \ref{fig:powerlaw} illustrates the power law fitting results of two CCF choices (VGG-16 conv3-3 and conv4-3) on two domains (faces and pedestrians). We determine whether power law holds based on the fitting deviation at nearby scales (i.e. small $\sigma$ at small scale values). In the domain of faces where images are of high-resolution and collected from the Internet, power law stands in conv3-3 (with $\sigma$ near $0$ when $log2(scale)$ is less than $-0.5$) but fails in conv4-4 (with $\sigma$ near $0.1$ when $log2(scale)$ is less than $-0.5$), as the latter feature representation is more sparse than the former. Experimental results in Section 4 further prove that power law holds in faces domain using conv3-3 features with negligible performance reduce. In the domain of pedestrians where VGA images are captured with a video camera, power law hardly fits both feature representations (with large $\sigma$) and we observe an $\sim3\%$ drop in performance on the above-mentioned evaluation protocol in practice. The reason for the failure of power law in pedestrian images is likely to be the degraded image quality.

\begin{figure}[t]
\begin{center}
   \includegraphics[width=0.9\linewidth]{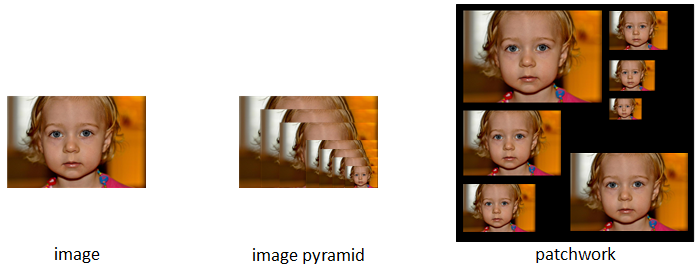}
\end{center}
   \caption{Illustration of how patchwork works.}
\label{fig:patchwork}
\end{figure}

\textbf{Patchwork:} 
The idea of patchwork was first proposed in~\cite{dubout2012exact} known as Bottom-Left Fill algorithm. Recent works on CNN framework also borrow the same idea~\cite{iandola2014densenet,papandreou2014untangling}. As the input image size of a CNN model is usually fixed due to implementation restrictions, the idea of patchwork is straightforward that images at different scales can be put together to form a large input image for feature extraction (as shown in Fig.~\ref{fig:patchwork}). The input image size of the CNN model is therefore set relatively large (say $1.5\times$ of average image size). Patchwork can get considerable speed boost in feature pyramid construction especially at dense scales, only if we handle the border effect between adjacent patches well. In our implementation, we add 16 pixel padding around each image when forming a large input image since the down-sampling factor of our CCF is 4. If the image is larger than the CNN input image size, we segment the image into small regions to fit into CNN input size rather than warp it. In doing so, our implementation can do patchwork with no approximation. Through experiments the patchwork can achieves $1.5\sim3$ times speed up compared with per-scale computation. The speed up factor depends on the implementation details and number of scales in feature pyramid.

\subsection{Model learning}

After handling with the feature extraction and fast computation, we now move to the inference and learning part. Generally we follow the learning pipelines used in channel features framework~\cite{dollar2014fast}. The learning workflow can be summarized as that ensemble of decision trees is learned in a boosting manner on the candidate features formed by pixel values in CCF. It is noted that for different tasks, the candidate features can have different forms, like single pixel lookups for pedestrian detection and pair-wise difference of two pixels for edge detection. While in inference stage, decision tree model is applied on dense image patches and output of each decision tree is accumulated to get the final result, during which only two operations are needed, namely pixel lookup and value comparison. Compared with CNN model, the boosting forest model is extremely light-weight with only tens of thousands parameters, easy to learn with fewer hyper-parameters, and less prone to over-fitting and adversarial examples~\cite{goodfellow2014explaining,szegedy2013intriguing}. In the following part we introduce in detail the models we use in different tasks.

\textbf{Pedestrian and face detection:} Decision trees are learned directly on pixel lookups of the given CNN feature maps like the way in~\cite{dollar2014fast,yang2014aggregate}. Since the feature representation is both sparse and robust due to multiple layers of no-linear transformation, no pre or post smoothing is needed. Similar to channel features, randomness is also beneficial in learning models on CCF for better efficiency in time and memory as well as better generalization ability.

\textbf{Edge detection and proposal generation:} For edge detection we deploy the structural learning approach in~\cite{dollar2014fastedge}. Model is built on $32\times32$ image patches, and candidate features are formed with pixel lookups and pairwise differences. We up-sample each image patch to keep the output feature map dimension as $16\times16$. Feature maps are smoothed with 2-pixel radius and 8-pixel radius each to generate pixel lookups and pairwise differences respectively. The learning paradigm remains the same as \cite{dollar2014fastedge}. As for the proposal generation task, we inherit the algorithm introduced in \cite{zitnick2014edge}, which is based on the results of edge detection.

\section{Experiments}

One reason that we select the low-level CNN features is that it is a good balance between feature representativeness and generalization ability. Low-level features are just not so abstract to become task-specific, nor so basic to be sensitive to appearance variations. We verify this on four different tasks, ranging from low-level to high-level vision, which are edge detection, object proposal generation, pedestrian and face detection. On the benchmark of each task, the proposed approach achieves state-of-the-art performances and outperforms hand-crafted filtered channel features and CNN based methods.
 
\subsection{Implementation details}

We use `conv3-3' layer in VGG-16 model as final choice of CCF representation, as it performs best in all tasks and with which the power law holds in the domain of faces. We implement feature extraction of CCF with Caffe toolkit~\cite{jia2014caffe}. Sliding window approach is deployed for all tasks during testing. Patchwork is used in CCF pyramid construction, where we use an input image size of $932\times932$ and set 6 scales per octave. Power law based approximation in nearby scales is used in face detection only (see Section 3.2 for reasons). 

\subsection{Feature analysis}

In this section we want to analyze the correlation between number of feature maps and the performance in the family of channel features. We use an imperfect but straightforward way to determine how much significance each feature map accounts for in the forest model by counting its occurrence in all forest nodes. As CCF feature representation is made up of 256 feature maps, we first train a boosting forest model using all feature maps, and choose the top 10, 40 and 128 most occurred (most selected) maps in the forest, and use these maps only to train new models, named as CCF-10, CCF-40 and CCF-128.

\begin{figure}[t]
\begin{center}
   \includegraphics[width=0.85\linewidth]{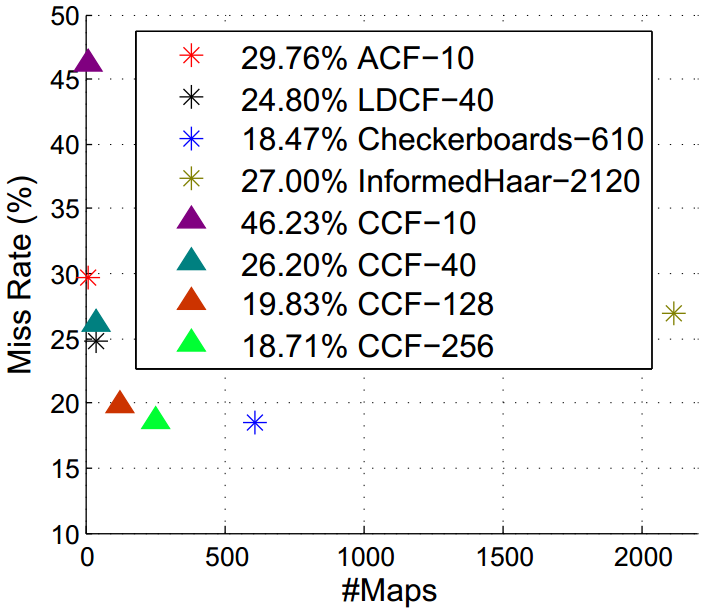}
\end{center}
   \caption{Performances of variants of channel features with different \#maps. Best view in color.}
\label{fig:analysis}
\end{figure}

Seeing from Table \ref{tab:occur} and Fig.~\ref{fig:analysis}, we get an interesting observation. As \#maps increases, the occurrence shows a tendency of dominance by the minority, and this tendency seems to be good for better performance, as the representative information is distributed non-uniformly across maps. The representative power lies in the form of combination of different maps, especially maps with different discriminativeness. The relatively poor performance of CCF-10 and CCF-40 compared with ACF-10 and LDCF-40 proves this supposition, as greedily selected maps are all very discriminative and they may share similar visual clues, therefore adding them up brings little boost in performance. How to choose a combination of limited number of feature maps and achieve the best performance is an open question.

\begin{table}
\begin{center}
\begin{tabular}{|l|c|c|c|}
\hline
Method-\#maps &N=10&N=20&N=50\\
 \hline\hline
 ACF-10 & 12\% & 23\% & 56\% \\
 LDCF-40 & 17\% & 30\% & 62\% \\
 \hline
 CCF-10 & 11\% & 22\% & 54\% \\
 CCF-40 & 13\% & 26\% & 56\% \\
 CCF-128 & 16\% & 29\% & 59\% \\
 CCF-256 & \textbf{22\%} & \textbf{36\%} & \textbf{70\%} \\
 \hline
 \end{tabular}
\end{center}
\caption{Occurrence of top N\% most occurred maps. As \#maps increases, the occurrence shows a tendency of long-tail distribution.}
\label{tab:occur}
\end{table}

\subsection{Pedestrian detection}

Caltech pedestrian benchmark~\cite{Dollar2012PAMI} is used for training and testing. Positive data is collected at a sampling interval of 4 frames from videos, while negative samples are collected by training a baseline aggregate channel features detector with 3 rounds of hard negative mining. We set model size to $128\times64$ for feature extraction and up-sample the image with a factor of 2 in feature pyramid construction for detecting pedestrians taller than 50 pixels. We make a variant of CCF by adding 10 HOG+LUV channels to CCF, called CCF+CF. The results are shown in Table ~\ref{tab:ped}.

We set new record (17.32\% by CCF+CF, slightly better than CCF) among all published methods without using temporal information, outperforming the best Checkerboards~\cite{Zhang2015Cvpr} with 1.15\% gain. Note that actually CCF has smaller feature dimension and less training data than Checkerboards (which has 610 feature maps and a sampling interval of 3 frames). CCF also has a great edge over CNN based methods JointDeep~\cite{ouyang2013joint}, SDN~\cite{Luo2014switchable} and AlexNet+ImageNet~\cite{Hosang2015Cvpr}, beating the best one with 6.00\% gain, while some of these CNN models are designed sophisticatedly. Note that there are still potentials in further improvements by replacing sliding window with region proposals (for example, the output of a weak pedestrian detector) or using more training data (at most 4 times than current setting) as well as hard negative mining. With all improvements deployed, we expect a further 3.00\% gain by tentative experimental results.

\begin{table}
\begin{center}
\begin{tabular}{|l|c|}
\hline
Method & Miss Rate (\%)\\
\hline\hline
JointDeep~\cite{ouyang2013joint} & +22.00\\
SDN~\cite{Luo2014switchable} & +20.55\\
InformedHaar~\cite{zhang2014informed} & +17.28\\
ACF-Caltech+~\cite{nam2014local} & +9.44\\
LDCF~\cite{nam2014local} & +7.48\\
AlexNet+ImageNet~\cite{Hosang2015Cvpr} & +6.00\\
Katamari (opt)~\cite{Benenson2014Eccvw} & +5.17\\
SpatialPooling+ (opt)~\cite{paisitkriangkrai2014strengthening} & +4.57\\
Checkerboards~\cite{Zhang2015Cvpr} & +1.15\\
\hline
CCF & +1.39\\
CCF+CF & \textbf{17.32}\\
\hline
\end{tabular}
\end{center}
\caption{Evaluation results of pedestrian detection on Caltech Pedestrian Benchmark under reasonable setting. Results of other methods are shown with relative performance gap. Methods using optical flow are marked with `opt'.}
\label{tab:ped}
\end{table}

\begin{figure}[t]
\begin{center}
  \includegraphics[width=.95\linewidth]{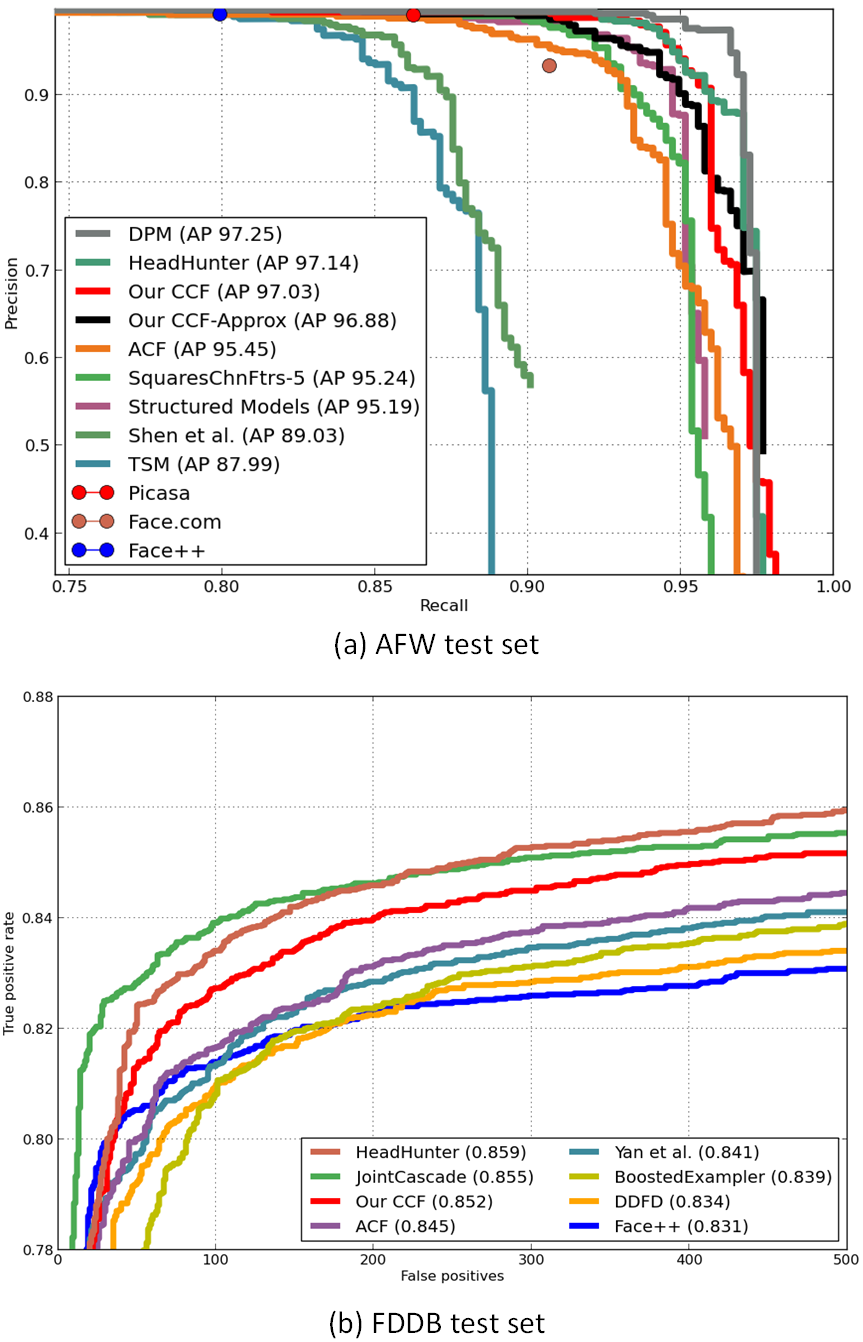}
\end{center}
   \caption{Evaluation results of face detection on (a) AFW and (b) FDDB test sets. Curves of proposed method are marked in red and black. Best viewed in color (zoom in).}
\label{fig:detection}
\end{figure}

\subsection{Face detection}

For face detection, we use the same training data as \cite{yang2014aggregate} and similar boosting paradigm (the depth of decision tree is changed from 2 to 3). An interesting observation is that with CCF feature representation, we can now handle multiple views with one boosting model instead of training multi-view models as in~\cite{yang2014aggregate,mathias2014face}. We evaluate the CCF face detector on two popular face detection test sets, AFW~\cite{zhu2012face} and FDDB~\cite{fddbTech}. We use the evaluation toolbox provided by~\cite{mathias2014face} to produce the curves, shown in Fig.~\ref{fig:detection}.

In AFW, CCF surpasses the baseline ACF in AP value with a 1.5\% gain, while is competitive with state-of-the-arts DPM~\cite{mathias2014face} and HeadHunter~\cite{mathias2014face}. However, CCF achieves the highest maximum recall rate among all algorithms. The deficiency in accuracy is mainly due to badly located bounding boxes. In addition, we also verify the power law feature approximation technique (described in section 3.2) in face detection (`Our CCF-Approx'), showing that the approximation leads to tolerant performance drop, while bringing a $2.4\times$ speed up during testing with patchwork deployed and $4\times$ speed up without patchwork.

In FDDB, it is no surprise that CCF beats the baseline ACF method~\cite{yang2014aggregate}. What should be noted is that CCF considerably outperforms another CNN based method DDFD \cite{farfade2015multi} which fine-tunes the CNN model to face detection, showing the effectiveness of boosting forest as a way to apply pre-trained CNN models to specific domains. The two state-of-the-art methods HeadHunter~\cite{mathias2014face} and JointCascade~\cite{chen2014joint} get better results for a reason. HeadHunter trains multi-scale and multi-view models and learns a transformation from rectangular outputs to ellipses while CCF trains one model and doesn't learn the transformation. JointCascade uses facial landmark annotations and adopts an additional post classifier to largely eliminate false positives, which are both orthogonal to our work.

\subsection{Edge detection}

We deploy the structured forests approach SE~\cite{dollar2014fastedge} for edge detection. For the feature representation part, as the feature down-sampling factor in SE is 2, while CCF has a down-sampling factor of 4, we up-sample the input image by a factor of 2. Also, SE uses multi-scale gradient channels, while we observe no improvements in using multi-scale CCF, therefore we just use the single-scale one. We train and evaluate the algorithm on BSDS500 dataset. From the results in Table ~\ref{tab:edge}, we can see that CCF outperforms SE with single-scale detection in all three metrics. When multi-scale detection (marked as `+ms' in the table) is used, SE gets a considerable improvement while CCF improves only a little. As for comparison with other state-of-the-art algorithms, different algorithms seem to have advantages in different metrics. Specifically, CCF beats other state-of-the-arts in AP value. What's more, adding 10 HOG+LUV channels to CCF feature representation (marked as `+CF' in the table) improves the ODS and OIS metrics a little but degrades the AP value.

\begin{table}
\begin{center}
\begin{tabular}{|l|c|c|c|}
\hline
Method & ODS & OIS & AP\\
\hline\hline
Human & 0.80 & 0.80 & -\\
\hline
DeepNet~\cite{kivinen2014visual} & 0.738 & 0.759 & 0.758\\
SE~\cite{dollar2014fastedge} & 0.739 & 0.759 & 0.792\\
SE+ms~\cite{dollar2014fastedge} & 0.746 & 0.767 & 0.803\\
MCG~\cite{arbelaez2014multiscale} & 0.747 & \textbf{0.779} & 0.759\\
DeepEdge~\cite{bertasius2014deepedge} & \textbf{0.753} & 0.772 & 0.807\\
\hline
CCF & 0.741 & 0.761 & 0.808\\
CCF+ms & 0.744 & 0.767 & \textbf{0.809}\\
CCF+ms+CF & 0.745 & 0.768 & 0.807\\
\hline
\end{tabular}
\end{center}
\caption{Evaluation results of edge detection on BSDS500 dataset. Three standard metrics are used, which are fixed contour threshold (ODS), per-image best threshold (OIS), and average precision (AP).}
\label{tab:edge}
\end{table}

\begin{table}
\newcommand{\tabincell}[2]{\begin{tabular}{@{}#1@{}}#2\end{tabular}}
\begin{center}
\begin{tabular}{|l|c|c|c|c|c|}
\hline
Methods & AUC & N@50\% & N@75\% & Recall \\
\hline\hline
BING~\cite{cheng2014bing} & 0.20 & - & - & 29\%\\
Objectness~\cite{alexe2010object}  & 0.27 & 584 & - & 68\%\\
Sel. Search~\cite{uijlings2013selective}  & 0.40 &199 & 1434 & 87\%\\
CPMC~\cite{carreira2010constrained}  & 0.41 & 111 & - & 65\%\\
EdgeBoxes~\cite{zitnick2014edge}  & 0.46 & 108 & 800 & 87\%\\
\hline
CCF & \textbf{0.48} & \textbf{89} & \textbf{649} & \textbf{88\%}\\
\hline
\end{tabular}
\end{center}
\caption{Evaluation results of object proposal generation on PASCAL VOC 2007 test set with IoU threshold of 0.7. Metrics are Area Under Curve (AUC), number of proposals needed to reach 50\% and 75\% recall and maximum recall rate. }
\label{tab:edgebox}
\end{table}

\subsection{Object proposal generation}

As the results in edge detection are mixed and we could not draw a convincing conclusion, here we adopt the EdgeBoxes approach \cite{zitnick2014edge} to generate object proposals directly from from edge detection results above. For efficiency, we use a single-scale version of CCF based edge detector. We evaluate the CCF based algorithm on the PASCAL VOC 2007 test set~\cite{everingham2010pascal}, which has well-annotated object bounding boxes. The results are shown in Table ~\ref{tab:edgebox}. Compared with other state-of-the-art algorithms, CCF achieves the best result in all four metrics, which are Area Under Curve, number of proposals needed to reach 50\% and 75\% recall and maximum recall rate. The results show that the edges detected by CCF are better than original channel features approach, at least with regard to forming object proposals.

\subsection{Speed}

By definition, CCF is apparently slower than channel features approach, and faster than end-to-end CNN methods (e.g. R-CNN) as high level connections in CNN take up roughly half the whole computation time~\cite{donahue2013decaf}. To be specific and accurate, we compare the speed of CCF with channel features variants and a popular CNN approach R-CNN in the task of pedestrian detection. The current implementation suffers from the sliding window mechanism, while some other detectors use object proposals as input due to heavy computation burden. Therefore, we evaluate speeds both on whole image detection and on one input window, in order to remove the effect of different inputs. Note that CCF is also applicable to object proposals as input, which is very likely to increase current performance in both detection speed and accuracy.

Results are shown in Table ~\ref{tab:speed}, which are in accord with our intuition. Note that due to a larger model size, ACF and CCF up-scale the image to 2 times in order to detect pedestrians taller than 50 pixels, so they are actually computing feature pyramid of a $1280\times960$ image. From the specific numbers, although CCF runs slowly on whole image, the time cost on single window is relatively low. Plus the fact that acceleration techniques used in Fast R-CNN~\cite{girshick2015fast} are also applicable to CCF, CCF has the potential to detect objects at real time speed.

\begin{table}
\begin{center}
\begin{tabular}{|l|c|c|}
\hline
Method & T(image)/s & T(window)/s\\
\hline\hline
ACF & 0.6 & \textless 0.01\\
LDCF  & 0.6 & \textless 0.01\\
R-CNN (CaffeNet) & 12.0 & 0.02\\
R-CNN (VGG-16) & 47.0$^\dagger$ & - \\
Fast R-CNN (VGG-16) & 0.4 & 0.02\\
\hline
CCF & 13.0 & \multirow{3}{*}{0.01}\\
CCF+PW & 8.6 &\\
CCF+PW+PL & 3.7 &\\
\hline
\end{tabular}
\end{center}
\caption{Speed comparison of CCF, ACF, LDCF, R-CNN and Fast R-CNN in pedestrian detection. Measurements are time costs on whole image and one window respectively, timed from image/window input to output boxes/score. The image is $640\times480$ large while the window is $128\times64$ large. `ACF' and `LDCF' are run on CPU while others are run on GPU. $^\dagger$Number referred from \cite{girshick2015fast}. The time of whole image detection of three R-CNN variants all exclude the time cost of region proposal generation. `PW' means patchwork while `PL' means power law.}
\label{tab:speed}
\end{table}

\section{Conclusion}
In this paper, we revisit the popular channel features approach and Convolutional Neural Networks approach, and propose an integrated method called Convolutional Channel Features (CCF) by combining the low-level CNN features and boosting forest model together. CCF benefits from the rich representative capacity of CNN to guarantee outstanding performance in various vision tasks, as well as the efficiency in inference and learning from boosting forest model. The proposed method achieves state-of-the-art performances in edge detection, object proposal generation, pedestrian and face detection, showing potentials for use in mobile and embedded devices with small model size.

\section{Acknowledgement}
We gratefully thank reviewers for their valuable advice and NVIDIA for the GPU donation. This work was supported by the Chinese National Natural Science Foundation Projects \#61203267, \#61375037, \#61473291, \#61572501, \#61572536, National Science and Technology Support Program Project \#2013BAK02B01, Chinese Academy of Sciences Project No. KGZD-EW-102-2, and AuthenMetric R\&D Funds.

{\footnotesize
\bibliographystyle{ieee}
\bibliography{egbib}

\begin{thebibliography}{10}\itemsep=-1pt

\bibitem{alexe2010object}
B.~Alexe, T.~Deselaers, and V.~Ferrari.
\newblock What is an object?
\newblock In {\em Computer Vision and Pattern Recognition (CVPR), 2010 IEEE
  Conference on}, pages 73--80. IEEE, 2010.

\bibitem{arbelaez2014multiscale}
P.~Arbelaez, J.~Pont-Tuset, J.~Barron, F.~Marques, and J.~Malik.
\newblock Multiscale combinatorial grouping.
\newblock In {\em Computer Vision and Pattern Recognition (CVPR), 2014 IEEE
  Conference on}, pages 328--335. IEEE, 2014.

\bibitem{benenson2013seeking}
R.~Benenson, M.~Mathias, T.~Tuytelaars, and L.~Van~Gool.
\newblock Seeking the strongest rigid detector.
\newblock In {\em CVPR}. IEEE, 2013.

\bibitem{Benenson2014Eccvw}
R.~Benenson, M.~Omran, J.~Hosang, and B.~Schiele.
\newblock Ten years of pedestrian detection, what have we learned?
\newblock In {\em Computer Vision-ECCV 2014 Workshops}, pages 613--627.
  Springer, 2014.

\bibitem{bertasius2014deepedge}
G.~Bertasius, J.~Shi, and L.~Torresani.
\newblock Deepedge: A multi-scale bifurcated deep network for top-down contour
  detection.
\newblock {\em arXiv preprint arXiv:1412.1123}, 2014.

\bibitem{carreira2010constrained}
J.~Carreira and C.~Sminchisescu.
\newblock Constrained parametric min-cuts for automatic object segmentation.
\newblock In {\em Computer Vision and Pattern Recognition (CVPR), 2010 IEEE
  Conference on}, pages 3241--3248. IEEE, 2010.

\bibitem{chen2014joint}
D.~Chen, S.~Ren, Y.~Wei, X.~Cao, and J.~Sun.
\newblock Joint cascade face detection and alignment.
\newblock In {\em Computer Vision--ECCV 2014}, pages 109--122. Springer, 2014.

\bibitem{cheng2014bing}
M.-M. Cheng, Z.~Zhang, W.-Y. Lin, and P.~Torr.
\newblock Bing: Binarized normed gradients for objectness estimation at 300fps.
\newblock In {\em Computer Vision and Pattern Recognition (CVPR), 2014 IEEE
  Conference on}, pages 3286--3293. IEEE, 2014.

\bibitem{dalal2005histograms}
N.~Dalal and B.~Triggs.
\newblock Histograms of oriented gradients for human detection.
\newblock In {\em CVPR}. IEEE, 2005.

\bibitem{PMT}
P.~Doll\'ar.
\newblock {P}iotr's {C}omputer {V}ision {M}atlab {T}oolbox ({PMT}).
\newblock \url{http://vision.ucsd.edu/~pdollar/toolbox/doc/index.html}.

\bibitem{dollar2014fast}
P.~Doll{\'a}r, R.~Appel, S.~Belongie, and P.~Perona.
\newblock Fast feature pyramids for object detection.
\newblock {\em PAMI}, 2014.

\bibitem{ICF}
P.~Doll\'ar, Z.~Tu, P.~Perona, and S.~Belongie.
\newblock Integral channel features.
\newblock In {\em BMVC}, 2009.

\bibitem{DollarCVPR10pose}
P.~Doll\'ar, P.~Welinder, and P.~Perona.
\newblock Cascaded pose regression.
\newblock In {\em CVPR}, 2010.

\bibitem{Dollar2012PAMI}
P.~Doll\'ar, C.~Wojek, B.~Schiele, and P.~Perona.
\newblock Pedestrian detection: An evaluation of the state of the art.
\newblock {\em PAMI}, 34, 2012.

\bibitem{dollar2014fastedge}
P.~Doll{\'a}r and C.~Zitnick.
\newblock Fast edge detection using structured forests.
\newblock {\em PAMI}, 2014.

\bibitem{donahue2013decaf}
J.~Donahue, Y.~Jia, O.~Vinyals, J.~Hoffman, N.~Zhang, E.~Tzeng, and T.~Darrell.
\newblock Decaf: A deep convolutional activation feature for generic visual
  recognition.
\newblock {\em arXiv preprint arXiv:1310.1531}, 2013.

\bibitem{dubout2012exact}
C.~Dubout and F.~Fleuret.
\newblock Exact acceleration of linear object detectors.
\newblock In {\em Computer Vision--ECCV 2012}, pages 301--311. Springer, 2012.

\bibitem{everingham2010pascal}
M.~Everingham, L.~Van~Gool, C.~K. Williams, J.~Winn, and A.~Zisserman.
\newblock The pascal visual object classes (voc) challenge.
\newblock {\em International journal of computer vision}, 88(2):303--338, 2010.

\bibitem{farfade2015multi}
S.~S. Farfade, M.~Saberian, and L.-J. Li.
\newblock Multi-view face detection using deep convolutional neural networks.
\newblock {\em arXiv preprint arXiv:1502.02766}, 2015.

\bibitem{girshick2015fast}
R.~Girshick.
\newblock Fast r-cnn.
\newblock {\em arXiv preprint arXiv:1504.08083}, 2015.

\bibitem{girshick2014rich}
R.~Girshick, J.~Donahue, T.~Darrell, and J.~Malik.
\newblock Rich feature hierarchies for accurate object detection and semantic
  segmentation.
\newblock In {\em CVPR}. IEEE, 2014.

\bibitem{girshick2014deformable}
R.~Girshick, F.~Iandola, T.~Darrell, and J.~Malik.
\newblock Deformable part models are convolutional neural networks.
\newblock {\em arXiv preprint arXiv:1409.5403}, 2014.

\bibitem{gkioxari2014r}
G.~Gkioxari, B.~Hariharan, R.~Girshick, and J.~Malik.
\newblock R-cnns for pose estimation and action detection.
\newblock {\em arXiv preprint arXiv:1406.5212}, 2014.

\bibitem{goodfellow2014explaining}
I.~J. Goodfellow, J.~Shlens, and C.~Szegedy.
\newblock Explaining and harnessing adversarial examples.
\newblock {\em arXiv preprint arXiv:1412.6572}, 2014.

\bibitem{gupta2014learning}
S.~Gupta, R.~Girshick, P.~Arbel{\'a}ez, and J.~Malik.
\newblock Learning rich features from rgb-d images for object detection and
  segmentation.
\newblock In {\em ECCV}. Springer, 2014.

\bibitem{he2014spatial}
K.~He, X.~Zhang, S.~Ren, and J.~Sun.
\newblock Spatial pyramid pooling in deep convolutional networks for visual
  recognition.
\newblock In {\em Computer Vision--ECCV 2014}, pages 346--361. Springer, 2014.

\bibitem{Hosang2015Cvpr}
J.~Hosang, M.~Omran, R.~Benenson, and B.~Schiele.
\newblock Taking a deeper look at pedestrians.
\newblock In {\em CVPR}, 2015.

\bibitem{iandola2014densenet}
F.~Iandola, M.~Moskewicz, S.~Karayev, R.~Girshick, T.~Darrell, and K.~Keutzer.
\newblock Densenet: Implementing efficient convnet descriptor pyramids.
\newblock {\em arXiv preprint arXiv:1404.1869}, 2014.

\bibitem{fddbTech}
V.~Jain and E.~Learned-Miller.
\newblock Fddb: A benchmark for face detection in unconstrained settings.
\newblock Technical Report UM-CS-2010-009, University of Massachusetts,
  Amherst, 2010.

\bibitem{jia2014caffe}
Y.~Jia, E.~Shelhamer, J.~Donahue, S.~Karayev, J.~Long, R.~Girshick,
  S.~Guadarrama, and T.~Darrell.
\newblock Caffe: Convolutional architecture for fast feature embedding.
\newblock In {\em Proceedings of the ACM International Conference on
  Multimedia}, pages 675--678. ACM, 2014.

\bibitem{karianakis2015boosting}
N.~Karianakis, T.~J. Fuchs, and S.~Soatto.
\newblock Boosting convolutional features for robust object proposals.
\newblock {\em arXiv preprint arXiv:1503.06350}, 2015.

\bibitem{kivinen2014visual}
J.~J. Kivinen, C.~K. Williams, N.~Heess, and D.~Technologies.
\newblock Visual boundary prediction: A deep neural prediction network and
  quality dissection.
\newblock In {\em AISTATS}, volume~1, page~9, 2014.

\bibitem{krizhevsky2012imagenet}
A.~Krizhevsky, I.~Sutskever, and G.~E. Hinton.
\newblock Imagenet classification with deep convolutional neural networks.
\newblock In {\em Advances in neural information processing systems}, pages
  1097--1105, 2012.

\bibitem{lecun1998gradient}
Y.~LeCun, L.~Bottou, Y.~Bengio, and P.~Haffner.
\newblock Gradient-based learning applied to document recognition.
\newblock {\em Proceedings of the IEEE}, 86(11):2278--2324, 1998.

\bibitem{lin2013nin}
M.~Lin, Q.~Chen, and S.~Yan.
\newblock Network in network.
\newblock {\em CoRR}, abs/1312.4400, 2013.

\bibitem{long2014fully}
J.~Long, E.~Shelhamer, and T.~Darrell.
\newblock Fully convolutional networks for semantic segmentation.
\newblock {\em arXiv preprint arXiv:1411.4038}, 2014.

\bibitem{Luo2014switchable}
P.~Luo, Y.~Tian, X.~Wang, and X.~Tang.
\newblock Switchable deep network for pedestrian detection.
\newblock In {\em Computer Vision and Pattern Recognition (CVPR), 2014 IEEE
  Conference on}, pages 899--906. IEEE, 2014.

\bibitem{mathias2014face}
M.~Mathias, R.~Benenson, M.~Pedersoli, and L.~Van~Gool.
\newblock Face detection without bells and whistles.
\newblock In {\em Computer Vision--ECCV 2014}, pages 720--735. Springer, 2014.

\bibitem{nam2014local}
W.~Nam, P.~Doll{\'a}r, and J.~H. Han.
\newblock Local decorrelation for improved pedestrian detection.
\newblock In {\em Advances in Neural Information Processing Systems}, pages
  424--432, 2014.

\bibitem{oquab2014learning}
M.~Oquab, L.~Bottou, I.~Laptev, and J.~Sivic.
\newblock Learning and transferring mid-level image representations using
  convolutional neural networks.
\newblock In {\em Computer Vision and Pattern Recognition (CVPR), 2014 IEEE
  Conference on}, pages 1717--1724. IEEE, 2014.

\bibitem{ouyang2013joint}
W.~Ouyang and X.~Wang.
\newblock Joint deep learning for pedestrian detection.
\newblock In {\em Computer Vision (ICCV), 2013 IEEE International Conference
  on}, pages 2056--2063. IEEE, 2013.

\bibitem{paisitkriangkrai2014strengthening}
S.~Paisitkriangkrai, C.~Shen, and A.~van~den Hengel.
\newblock Strengthening the effectiveness of pedestrian detection with
  spatially pooled features.
\newblock In {\em Computer Vision--ECCV 2014}, pages 546--561. Springer, 2014.

\bibitem{papandreou2014untangling}
G.~Papandreou, I.~Kokkinos, and P.-A. Savalle.
\newblock Untangling local and global deformations in deep convolutional
  networks for image classification and sliding window detection.
\newblock {\em arXiv preprint arXiv:1412.0296}, 2014.

\bibitem{savalle8deformable}
P.-A. Savalle, S.~Tsogkas, G.~Papandreou, and I.~Kokkinos.
\newblock Deformable part models with cnn features.
\newblock In {\em 3rd Parts and Attributes Workshop, ECCV}, volume~8.

\bibitem{simonyan2014very}
K.~Simonyan and A.~Zisserman.
\newblock Very deep convolutional networks for large-scale image recognition.
\newblock {\em arXiv preprint arXiv:1409.1556}, 2014.

\bibitem{szegedy2014going}
C.~Szegedy, W.~Liu, Y.~Jia, P.~Sermanet, S.~Reed, D.~Anguelov, D.~Erhan,
  V.~Vanhoucke, and A.~Rabinovich.
\newblock Going deeper with convolutions.
\newblock {\em arXiv preprint arXiv:1409.4842}, 2014.

\bibitem{szegedy2013intriguing}
C.~Szegedy, W.~Zaremba, I.~Sutskever, J.~Bruna, D.~Erhan, I.~Goodfellow, and
  R.~Fergus.
\newblock Intriguing properties of neural networks.
\newblock {\em arXiv preprint arXiv:1312.6199}, 2013.

\bibitem{uijlings2013selective}
J.~R. Uijlings, K.~E. van~de Sande, T.~Gevers, and A.~W. Smeulders.
\newblock Selective search for object recognition.
\newblock {\em International journal of computer vision}, 104(2):154--171,
  2013.

\bibitem{viola2004robust}
P.~Viola and M.~J. Jones.
\newblock Robust real-time face detection.
\newblock {\em IJCV}, 2004.

\bibitem{yang2014aggregate}
B.~Yang, J.~Yan, Z.~Lei, and S.~Z. Li.
\newblock Aggregate channel features for multi-view face detection.
\newblock In {\em IJCB}. IEEE, 2014.

\bibitem{zhang2014part}
N.~Zhang, J.~Donahue, R.~Girshick, and T.~Darrell.
\newblock Part-based r-cnns for fine-grained category detection.
\newblock In {\em ECCV}. Springer, 2014.

\bibitem{zhang2014informed}
S.~Zhang, C.~Bauckhage, and A.~B. Cremers.
\newblock Informed haar-like features improve pedestrian detection.
\newblock In {\em Computer Vision and Pattern Recognition (CVPR), 2014 IEEE
  Conference on}, pages 947--954. IEEE, 2014.

\bibitem{Zhang2015Cvpr}
S.~Zhang, R.~Benenson, and B.~Schiele.
\newblock Filtered channel features for pedestrian detection.
\newblock In {\em CVPR}, 2015.

\bibitem{zhu2012face}
X.~Zhu and D.~Ramanan.
\newblock Face detection, pose estimation, and landmark localization in the
  wild.
\newblock In {\em Computer Vision and Pattern Recognition (CVPR), 2012 IEEE
  Conference on}, pages 2879--2886. IEEE, 2012.

\bibitem{zitnick2014edge}
C.~L. Zitnick and P.~Doll{\'a}r.
\newblock Edge boxes: Locating object proposals from edges.
\newblock In {\em ECCV}. Springer, 2014.

\bibitem{ZitnickDollarECCV14edgeBoxes}
C.~L. Zitnick and P.~Doll\'ar.
\newblock Edge boxes: Locating object proposals from edges.
\newblock In {\em ECCV}, 2014.

\bibitem{zou2014generic}
W.~Y. Zou, X.~Wang, M.~Sun, and Y.~Lin.
\newblock Generic object detection with dense neural patterns and regionlets.
\newblock {\em arXiv preprint arXiv:1404.4316}, 2014.

\end{thebibliography}
}

\end{document}